\documentclass[twoside,twocolumn,9pt]{article}
\usepackage{extsizes}
\usepackage[super,sort&compress,comma]{natbib} 
\usepackage[version=3]{mhchem}
\usepackage[left=1.5cm, right=1.5cm, top=1.785cm, bottom=2.0cm]{geometry}
\usepackage{balance}
\usepackage{mathptmx}
\usepackage{sectsty}
\usepackage{graphicx} 
\usepackage{lastpage}
\usepackage[format=plain,justification=justified,singlelinecheck=false,font={stretch=1.125,small,sf},labelfont=bf,labelsep=space]{caption}
\usepackage{float}
\usepackage{fancyhdr}
\usepackage{fnpos}
\usepackage[english]{babel}
\addto{\captionsenglish}{%
  
}
\usepackage{array}
\usepackage{droidsans}
\usepackage{charter}
\usepackage[T1]{fontenc}
\usepackage[usenames,dvipsnames]{xcolor}
\usepackage{setspace}
\usepackage[compact]{titlesec}
\usepackage{hyperref}
\usepackage{amssymb, amsmath}
\usepackage{algorithm} 
\usepackage{algpseudocode}
\usepackage{bm}

\usepackage{pdfpages}

\usepackage{epstopdf}

\definecolor{cream}{RGB}{222,217,201}

\DeclareMathOperator*{\argmax}{argmax}
\DeclareMathOperator*{\argmin}{argmin}

\begin{document}

\pagestyle{fancy}
\thispagestyle{plain}
\fancypagestyle{plain}{
\renewcommand{\headrulewidth}{0pt}
}

\makeFNbottom
\makeatletter
\renewcommand\LARGE{\@setfontsize\LARGE{15pt}{17}}
\renewcommand\Large{\@setfontsize\Large{12pt}{14}}
\renewcommand\large{\@setfontsize\large{10pt}{12}}
\renewcommand\footnotesize{\@setfontsize\footnotesize{7pt}{10}}
\makeatother

\renewcommand{\thefootnote}{\fnsymbol{footnote}}
\renewcommand\footnoterule{\vspace*{1pt}%
\color{cream}\hrule width 3.5in height 0.4pt \color{black}\vspace*{5pt}} 
\setcounter{secnumdepth}{5}

\makeatletter 
\renewcommand\@biblabel[1]{#1}            
\renewcommand\@makefntext[1]%
{\noindent\makebox[0pt][r]{\@thefnmark\,}#1}
\makeatother 
\renewcommand{\figurename}{\small{Fig.}~}
\sectionfont{\sffamily\Large}
\subsectionfont{\normalsize}
\subsubsectionfont{\bf}
\setstretch{1.125} 
\setlength{\skip\footins}{0.8cm}
\setlength{\footnotesep}{0.25cm}
\setlength{\jot}{10pt}
\titlespacing*{\section}{0pt}{4pt}{4pt}
\titlespacing*{\subsection}{0pt}{15pt}{1pt}

\fancyfoot{}
\fancyfoot[LO,RE]{\vspace{-7.1pt}\includegraphics[height=9pt]{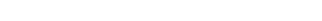}}
\fancyfoot[CO]{\vspace{-7.1pt}\hspace{13.2cm}\includegraphics{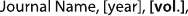}}
\fancyfoot[CE]{\vspace{-7.2pt}\hspace{-14.2cm}\includegraphics{head_foot/RF}}
\fancyfoot[RO]{\footnotesize{\sffamily{1--\pageref{LastPage} ~\textbar  \hspace{2pt}\thepage}}}
\fancyfoot[LE]{\footnotesize{\sffamily{\thepage~\textbar\hspace{3.45cm} 1--\pageref{LastPage}}}}
\fancyhead{}
\renewcommand{\headrulewidth}{0pt} 
\renewcommand{\footrulewidth}{0pt}
\setlength{\arrayrulewidth}{1pt}
\setlength{\columnsep}{6.5mm}
\setlength\bibsep{1pt}

\makeatletter 
\newlength{\figrulesep} 
\setlength{\figrulesep}{0.5\textfloatsep} 

\newcommand{\topfigrule}{\vspace*{-1pt}%
\noindent{\color{cream}\rule[-\figrulesep]{\columnwidth}{1.5pt}} }

\newcommand{\botfigrule}{\vspace*{-2pt}%
\noindent{\color{cream}\rule[\figrulesep]{\columnwidth}{1.5pt}} }

\newcommand{\dblfigrule}{\vspace*{-1pt}%
\noindent{\color{cream}\rule[-\figrulesep]{\textwidth}{1.5pt}} }

\makeatother

\twocolumn[
  \begin{@twocolumnfalse}
{\includegraphics[height=30pt]{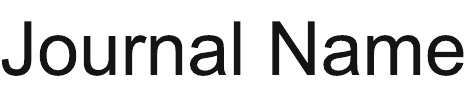}\hfill\raisebox{0pt}[0pt][0pt]{\includegraphics[height=55pt]{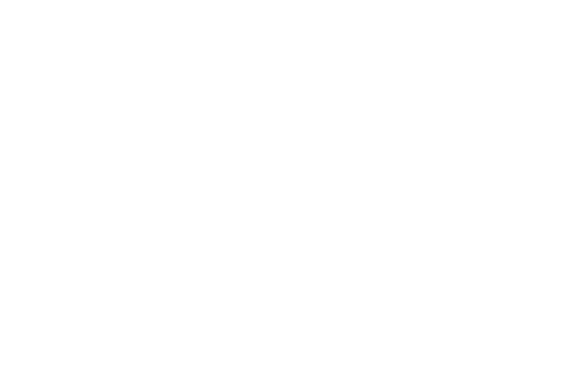}}\\[1ex]
\includegraphics[width=18.5cm]{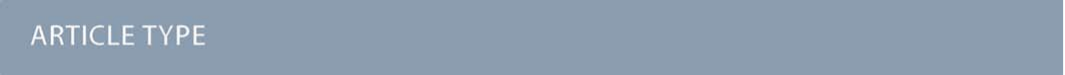}}\par
\vspace{1em}
\sffamily
\begin{tabular}{m{4.5cm} p{13.5cm} }

\includegraphics{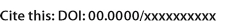} & \noindent\LARGE{\textbf{BEACON: A Bayesian Optimization Inspired Strategy for Efficient Novelty Search}} \\
\vspace{0.3cm} & \vspace{0.3cm} \\

 & \noindent\large{Wei-Ting Tang,\textit{$^{a, b}$} Ankush Chakrabarty,\textit{$^{c}$} and Joel A. Paulson$^{\ast}$\textit{$^{a, b}$}} \\

\includegraphics{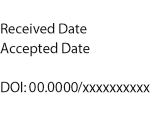} & \noindent\normalsize{Novelty search (NS) aims to uncover diverse system behaviors through simulation or experiment without requiring a pre-specified scalar objective. This capability is especially relevant to modern discovery problems in chemistry, materials science, and molecular design, where researchers often seek broad coverage of attainable property space rather than a single optimum and where each evaluation may require a costly computation or experiment. For such expensive black-box settings, we propose BEACON, a sample-efficient NS strategy inspired by Bayesian optimization principles. 
BEACON models the input-to-outcome mapping using multi-output Gaussian processes and selects new inputs by scoring how far plausible posterior outcomes lie from a denoised archive of previously observed outcomes. This gives a distance-based novelty acquisition that accounts for predictive uncertainty and observational noise while operating directly in continuous outcome space, rather than requiring direct optimization over a discretized partition of behaviors.
By leveraging efficient posterior sampling together with scalable high-dimensional Gaussian process models, the proposed framework can be extended to settings with large data sets and high-dimensional design variables. We demonstrate BEACON on established benchmark problems together with real-world case studies in materials and molecular discovery. Across these settings, BEACON consistently discovers broader sets of distinct behaviors than several competing baselines under limited evaluation budgets.
} \\

\end{tabular}

 \end{@twocolumnfalse} \vspace{0.6cm}

  ]

\renewcommand*\rmdefault{bch}\normalfont\upshape
\rmfamily
\section*{}
\vspace{-1cm}


\footnotetext{\textit{$^{a}$~University of Wisconsin-Madison, Department of Chemical and Biological Engineering, Madison, WI, USA.}}
\footnotetext{\textit{$^{b}$~The Ohio State University, Department of Chemical and Biomolecular Engineering, Columbus, OH, USA.}}
\footnotetext{\textit{$^{c}$~Insulet Corporation, Acton, MA, USA.}}
\footnotetext{\textit{$^{*}$~Correspondence: joel.paulson@wisc.edu}}




\section{Introduction}
\label{sec:introduction}

Across chemistry, materials science, and molecular design, researchers increasingly rely on expensive simulations, data-driven models, and automated experiments to probe complex black-box systems. In many of these settings, the goal is not only to identify a single best candidate, but also to uncover a \emph{diverse} set of attainable outcomes that can reveal new mechanisms, broaden training data, or expose promising but previously overlooked regions of design space. These needs arise naturally in applications such as molecular discovery, materials screening, and autonomous experimentation, where broad exploration of outcome space can be as valuable as objective-driven optimization~\cite{grizou2020curious,terayama2020pushing,stach2021autonomous,wang2022bayesian}.

To formalize this setting, consider a black-box map $f : X \mapsto O$ that takes user-defined system inputs $x \in X$ to outcomes (or behaviors) $f(x) \in O$. Classical derivative-free search methods, including simplex methods, meta-heuristics, and Bayesian optimization (BO), are designed primarily to \emph{optimize} some prescribed objective defined over the outcomes; see, for example,~\cite{rios2013derivative,shahriari2015taking,russell2016artificial,frazier2018tutorial}. This objective-centric viewpoint is highly effective when the user knows what should be maximized or minimized. However, if the real aim is to understand the range of outcomes that the system can realize, then reducing the problem to a single scalar objective can be restrictive and, in some cases, fundamentally misaligned with the scientific goal.

A simple example arises in molecular/materials discovery. Suppose $f$ maps a candidate design to a vector of experimentally relevant properties, such as solubility, uptake capacity, stability, or selectivity. If the objective is to identify \emph{new types} of property combinations rather than only the single best design under one fixed score, then standard optimization becomes awkward. Any scalarization will prioritize certain regions of the outcome space and can easily overlook scientifically useful intermediate or rare behaviors. More broadly, the corresponding search landscape is often highly multi-modal and deceptive, making exhaustive exploration prohibitively expensive when each evaluation involves a simulation, synthesis, or physical measurement.

Novelty search (NS) offers a natural alternative by shifting the emphasis from objective maximization to the discovery of previously unseen behaviors~\cite{lehman2011abandoning,lehman2011novelty}. Instead of rewarding progress toward a predefined target, NS promotes candidates whose outcomes differ from what has already been observed. In the classical archive-based view, a candidate is novel when its behavior is far from the archive of behaviors discovered so far. This is the sense in which we use the term NS throughout this work. The broader goal is to build a discovery campaign that covers a wide range of attainable behaviors under a limited evaluation budget. We later measure this coverage using reachability, but the acquisition function itself operates in the continuous outcome space rather than over a discrete coverage count. This distinction lets BEACON use a continuous, distance-based notion of novelty while still evaluating success in terms of behavior-space coverage.

This perspective has proven useful in automated scientific discovery~\cite{grizou2020curious,terayama2020pushing}, reinforcement learning~\cite{jackson2019novelty}, and more recently, autonomous experimentation workflows that explicitly seek previously unobserved phenomena rather than only improved objective values~\cite{bulanadi2025beyond}. Most established NS algorithms, however, rely on meta-heuristics such as evolutionary search. Genetic algorithms, for example, have been used to promote behavioral diversity in deceptive landscapes~\cite{gomes2015}; swarm-based methods have also been adapted for novelty-driven exploration~\cite{mouret2012}; and MAP-Elites~\cite{mouret2015} has become a widely used framework for illuminating trade-offs between diversity and performance. While powerful, these methods are typically sample-inefficient and therefore poorly matched to settings in which each function evaluation is costly.

In this work, we draw inspiration from BO to develop a \emph{sample-efficient} Bayesian novelty search strategy for expensive black-box systems. Rather than placing a surrogate model on a scalar objective, we model the full input-to-outcome relationship using Gaussian processes (GPs)~\cite{williams2006gaussian}. This distinction is important since the surrogate is used to reason directly about the outcomes a user wishes to explore, not to collapse them into a single utility function. As a result, the outcome space can be defined flexibly by the practitioner and may include multiple quantities of interest. In a drug discovery setting, for example, the outcome vector may encode efficacy, solubility, partitioning behavior, synthesizability, or stability, while in a materials setting it may encode multiple uptake or transport properties.

Our main contributions are summarized as follows: 
\begin{itemize}
    \item We introduce a new novelty search algorithm, referred to as \textbf{BEACON} (\textbf{B}ayesian \textbf{E}xploration \textbf{A}lgorithm for out\textbf{CO}me \textbf{N}ovelty), for noisy, expensive black-box systems. BEACON is designed to improve behavior-space reachability using a limited number of evaluations.
    \item We propose a Thompson sampling-based acquisition function for NS that scores candidate inputs by the distance between their sampled outcomes and a denoised archive of previously observed outcomes. This gives a continuous NS policy that accounts for predictive uncertainty and observation noise without requiring direct optimization over a fixed behavior partition.
    \item We extend BEACON to high-dimensional settings through both sparsity-inducing fully Bayesian priors and chemistry-aware surrogate modeling strategies.
    \item We demonstrate BEACON on a broad set of benchmark and real-world discovery problems, including metal-organic framework discovery for clean energy applications and molecular property discovery with input representations as large as 2133 dimensions.
\end{itemize}
Taken together, these results show that BEACON can substantially improve the efficiency of novelty-driven exploration in settings that are directly relevant to modern discovery workflows.

The rest of the paper is organized as follows. In Section~\ref{sec:prob-desc}, we define the black-box NS problem considered in this work. In Section~\ref{sec:background}, we present the required background on BO and multi-output GP regression and introduce key mathematical notation. Section~\ref{sec:beacon} introduces the proposed BEACON algorithm, including the acquisition function, implementation details, and extensions for high-dimensional problems and user-guided behavior constraints. Section~\ref{sec:results-discussion} presents benchmark comparisons and real-world case studies that illustrate the practical value of BEACON for sample-efficient discovery. Lastly, we provide some concluding remarks in Section \ref{sec:conclusions}

\section{Problem Formulation}
\label{sec:prob-desc}

We consider a vector-valued black-box function
\begin{align*}
    f:X\to O,
    \qquad
    f=\big(f^{(1)},\ldots,f^{(n)}\big),
\end{align*}
that maps an input space $X\subset \mathbb{R}^d$ to an $n$-dimensional outcome space $O\subset \mathbb{R}^n$. In this work, the outcomes are the behaviors we wish to explore. A behavior may be a scalar property, a vector of measured properties, a terminal state, a generated image, or another application-specific representation chosen by the practitioner. The key assumption is that this representation supports a meaningful notion of behavioral similarity, either through a distance metric, a finite partition, or both.

Although the examples in this paper mostly define behaviors using measured or predicted outcomes, the behavior representation can also include known input-derived quantities when they are scientifically relevant. For example, if two chemistries produce similar measured outcomes but differ in cost, synthesis route, reaction conditions, or material availability, these quantities can be included in the behavior representation (or in the distance metric used by the novelty score defined in Section~\ref{sec:beacon}). Known quantities do not require GP modeling; they can be appended directly to the sampled outcome representation when evaluating novelty.

We use the term ``discovery campaign'' to refer to the full sequence of evaluations made under the budget $T$. The campaign-level goal is to choose this sequence so that the observed outcomes cover as much of the attainable behavior space as possible, rather than optimizing a single scalar objective. To measure this coverage, we introduce a finite behavior space $B=\{1,\ldots,M\}$ together with behavior regions $\{C_b\}_{b\in B}$ that partition the relevant outcome space. The behavior assignment map $\phi:O\rightarrow B$ converts each continuous outcome to its behavior region index, so that $\phi(y)=b$ if $y\in C_b$.

At iteration $t=1,\ldots,T$, we query an input $x_t\in X$ and observe
\begin{align*}
    y_t=f(x_t)+\eta_t,\qquad \eta_t\sim \mathcal{N}(0,\sigma^2 I_n),
\end{align*}
where $\eta_t$ denotes independent Gaussian observation noise. Let
\begin{align*}
    I_t=\{\phi(f(x_i)):i=1,\ldots,t\}\subseteq B
\end{align*}
be the set of distinct behavior regions reached after $t$ evaluations. We write $|I_t|$ for the cardinality of this set, i.e., the number of distinct behavior-region indices reached by iteration $t$. We use reachability as one simple measure of campaign progress,
\begin{align}
    \mathrm{Reach}_t=\frac{|I_t|}{|B|},
    \label{eq:reachability}
\end{align}
where $|B|=M$ is the total number of behavior regions. Reachability is a bin-occupancy measure of coverage. It increases only when the campaign discovers a behavior region that has not been observed before. This makes it useful for novelty-driven discovery because it measures breadth of observed outcomes without requiring a trusted scalar objective or utility function. Other coverage metrics could also be used, depending on the application. For a loss-style convention, one could instead define the ``behavior gap'' as $1-\mathrm{Reach}_t$, which tracks the fraction of behavior regions that remain unreached and is analogous in spirit to regret-style measures used in bandit and BO settings. Here, we report reachability so that larger values consistently indicate broader coverage.

The corresponding sequential design problem is to choose a sampling policy that discovers as many distinct behavior regions as possible over the evaluation budget. One way to express this goal is through the following problem
\begin{align}
    \max_{\pi}\; \mathbb{E}_{\pi}\!\left[\sum_{t=1}^{T}\mathrm{Reach}_t\right],
    \label{eq:cumulative-reachability}
\end{align}
where $\pi$ denotes the sampling policy and the expectation is over any noise and policy randomness. This objective in \eqref{eq:cumulative-reachability} defines how we evaluate a discovery campaign, but it does not by itself specify the acquisition function used to choose the next input.

This distinction motivates the design of BEACON. A direct one-step reachability acquisition can be derived by maximizing the posterior probability that $f(x)$ lands in a behavior region not yet reached. Section~1 of the Supplementary Information derives this acquisition and provides a one-dimensional illustration comparing it with the BEACON novelty acquisition. This direct acquisition is natural when the behavior map $\phi$ is known in advance and the behavior space is small enough to enumerate. In many early-stage discovery settings, however, the practitioner may know how to represent and compare outcomes before knowing the right partition of outcome space into behaviors. Direct reachability acquisitions also require probability mass calculations over unreached behavior regions, which can become expensive or numerically delicate as the number of regions grows.

BEACON therefore uses NS in the archive-distance sense. Candidate outcomes are preferred when they are far from outcomes already observed. The acquisition operates directly in the continuous outcome space through a user-defined distance metric, while the finite behavior space $B$ is used to evaluate coverage after samples have been collected. This evaluation space can be constructed using a uniform grid, an $\varepsilon$-cover, clustering, or user-defined behavior regions. The choice sets the resolution of the reachability metric. A smaller $\varepsilon$ (or a finer partition more generally) distinguishes more outcomes and makes reachability stricter, while a coarser partition groups nearby outcomes together. Under the default BEACON acquisition, this choice affects the reported reachability and interpretation of the final discovered set, but not the sampling decisions themselves. We study the sensitivity of reported reachability to this behavior-space resolution in the Supplementary Information.

Because $f$ is expensive to evaluate, exhaustive exploration is generally impractical. We thus use a probabilistic surrogate for the input-to-outcome map. Specifically, BEACON places a multi-output Gaussian process (MOGP) prior on $f$. Given observations
\begin{align*}
    D_t=\{(x_i,y_i)\}_{i=1}^{t},
\end{align*}
the MOGP induces a posterior distribution $P(f\mid D_t)$ over the unknown vector-valued function. This posterior enables BEACON to denoise previously observed outcomes, quantify uncertainty in unobserved regions of input space, and guide sampling toward plausible new regions of outcome space. The framework permits correlated or independent output models; the experiments below use the modeling choices described in Section~3 and the Supplementary Information.

\section{Methodological Background}
\label{sec:background}

This section briefly reviews the two core ingredients underlying BEACON: multi-output Gaussian process (MOGP) surrogate modeling and Bayesian optimization (BO).

\subsection{Multi-Output Gaussian Process Surrogate Modeling}

Gaussian processes (GPs) provide a flexible nonparametric framework for modeling expensive black-box functions~\cite{williams2006gaussian}. A GP prior is fully specified by a mean function $\mu$ and a covariance (kernel) function $\kappa$. In BEACON, the surrogate is used to model the input-to-outcome map $f:X\rightarrow O$, where $f(x)=(f^{(1)}(x),\ldots,f^{(n)}(x))$ may contain one or more outcome coordinates.

The BEACON framework does not require a particular MOGP construction. In most of the experiments reported here, we use independent-output GP models, meaning that each outcome coordinate is modeled with its own scalar GP. This choice keeps the surrogate modeling component simple and makes the empirical results focus on the proposed NS policy rather than on advances in multi-output modeling. Correlated-output models could be substituted when cross-output structure is important, but this is not the focus of the present study.

For conciseness, it is useful to write the vector-valued surrogate using an output-indexed representation. Define
\[
h(x,j)=f^{(j)}(x),
\qquad
j\in J:=\{1,\ldots,n\},
\]
and place a scalar GP prior $h\sim GP(\mu,\kappa)$ over the extended input space $X\times J$; see, e.g., Kudva et al.~\cite{kudva2024robust}. A general kernel $\kappa((x,j),(x',j'))$ can encode correlations across outputs. In our experiments, however, we use the block-diagonal kernel
\begin{equation}
    \kappa\big((x,j),(x',j')\big)
    =
    \delta_{jj'}\,\kappa_j(x,x'),
    \label{eq:independent-output-kernel}
\end{equation}
where $\delta_{jj'}$ is the Kronecker delta and $\kappa_j$ is the scalar kernel for output $j$. Thus, \eqref{eq:independent-output-kernel} is equivalent to fitting one scalar GP per output coordinate.

Suppose we observe the indexed dataset $A=\{(x_i,j_i,y_i)\}_{i=1}^{N}$,
where $y_i$ is a noisy measurement of $h(x_i,j_i)$. Under Gaussian observation noise, the posterior remains a GP, $h\mid A\sim \mathcal{GP}(\mu_A,\kappa_A)$, with mean and covariance functions~\cite{williams2006gaussian,liu2018remarks}
\begin{subequations}
\label{eq:posterior-single-output}
\begin{align}
    &\mu_A(x,j) \\[-4mm]\notag
    &=
    \mu(x,j)
    +
    k_A(x,j)^\top
    \left(K_A+\sigma^2 I_N\right)^{-1}
    (y-\mu_A^{\mathrm{obs}}),
    \\
    &\kappa_A\big((x,j),(x',j')\big) \\[-4mm]\notag
    &=
    \kappa\big((x,j),(x',j')\big)
    -
    k_A(x,j)^\top
    \left(K_A+\sigma^2 I_N\right)^{-1}
    k_A(x',j').
\end{align}
\end{subequations}
Here, $y=[y_1,\ldots,y_N]^\top$, $\mu_A^{\mathrm{obs}}=[\mu(x_1,j_1),\ldots,\mu(x_N,j_N)]^\top$, $k_A(x,j)\in\mathbb{R}^N$ is the vector of covariance values between the test pair $(x,j)$ and the observed indexed inputs in $A$, and $K_A\in\mathbb{R}^{N\times N}$ is the covariance matrix over the observed indexed inputs.

With the independent-output kernel in \eqref{eq:independent-output-kernel}, these posterior equations decouple across outputs. The posterior mean vector and covariance matrix for $f(x)$ are then assembled from the scalar GP posteriors for each coordinate. This is the implementation used throughout the numerical studies (except for the MNIST case). In the chemistry and molecular examples, task-specific kernels are used for the relevant scalar GP components; these choices modify $\kappa_j$ but do not change the BEACON acquisition principle.

It is worth noting that, if output correlations are important, the independent-output GP can be replaced by a correlated multi-output kernel or by a structured high-dimensional output model. For example, in the MNIST case study we use a high-dimensional, multi-task GP model \cite{maddox2021bayesian} to model the full set of $14 \times 14$ pixels in the decoded image.

\subsection{Bayesian Optimization}

Bayesian optimization (BO) is a sequential strategy for optimizing an expensive black-box function. For exposition, consider the minimization problem
\[
x^\star=\argmin_{x\in X} f(x),
\]
where $X$ is the admissible search space and evaluations of $f$ may be noisy and costly. Classical BO alternates between two steps: fitting a probabilistic surrogate model to the available data and optimizing an acquisition function that quantifies the value of sampling a candidate point next.

Common acquisition functions include upper confidence bound (UCB)~\cite{srinivas2009gaussian}, expected improvement (EI)~\cite{ament2023unexpected}, and Thompson sampling (TS)~\cite{thompson1933likelihood}. TS is a randomized decision rule that draws a sample function $g$ from the surrogate posterior and then selects the optimizer of that sample. In the minimization setting, this corresponds to choosing $\argmin_{x\in X} g(x)$.
TS often performs well in practice and is especially attractive in our setting because it naturally balances exploration and exploitation while remaining easy to adapt beyond classical objective-based optimization~\cite{kandasamy2018parallelised}.

BEACON borrows this posterior sampling idea from BO but applies it to NS rather than scalar objective optimization. Instead of drawing a posterior sample to identify a likely optimizer of a scalar objective, BEACON draws a posterior sample of the input-to-outcome map and scores candidate inputs by how novel their sampled outcomes are relative to the denoised archive of previously observed outcomes. This adaptation is formally introduced in the next section.

\section{BEACON Method}
\label{sec:beacon}

We now present the proposed BEACON algorithm. We first introduce a Thompson sampling-based acquisition function for novelty search (NS), then summarize the overall algorithm, discuss its efficient optimization, and finally describe extensions for high-dimensional problems and user-guided behavior constraints.

\subsection{A Thompson Sampling Acquisition Function for Novelty Search}

Our goal is to discover novel system behaviors by exploring unobserved regions of the continuous outcome space $O$. A common strategy in the NS literature \cite{lehman2011abandoning} is to quantify how far a candidate outcome lies from previously observed outcomes. One standard novelty score is the average distance to the $k$ nearest neighbors in outcome space:
\begin{align}
\label{eq:novelty-metric}
\rho(x\mid D)
=
\frac{1}{k}
\sum_{i=1}^k
\mathrm{dist}\big(f(x), y_i^\star\big),
\end{align}
where $\{y_1^\star,\ldots,y_k^\star\}
\subset
\{y_1,\ldots,y_N\}$
are the $k$ nearest previously observed outcomes to $f(x)$ under the user-defined distance metric $\mathrm{dist}(\cdot)$ over $O$. 

Although intuitive, directly using \eqref{eq:novelty-metric} is problematic in our setting for two reasons. First, $f$ is unknown and expensive to query, so $\rho(x\mid D)$ cannot be evaluated without performing the experiment or simulation at $x$. Second, when observations are noisy, novelty scores based directly on raw outcomes can be distorted by measurement noise. In particular, it may occur that
\[
\phi\big(f(x)+\eta\big)
\neq
\phi\big(f(x)\big),
\]
leading to spurious behavior assignments.

To address these issues, we replace the true black-box map with its MOGP posterior surrogate $f\mid D
\sim
MOGP(\mu_{D},\kappa_{D})$. The surrogate serves two purposes. First, it denoises previously observed outcomes by replacing raw observations with posterior mean predictions. Second, it allows us to account for uncertainty in unobserved regions of the input space through posterior sampling. Using Thompson sampling, we draw a sample function $g(x) \sim f\mid D$
and define the BEACON acquisition function as
\begin{align}
\label{eq:ns-ts}
\alpha_{\mathrm{NS}}(x\mid g,D)
=
\frac{1}{k}
\sum_{i=1}^k
\mathrm{dist}\big(
g(x),
\mu_{D}(x_i^\star)
\big),
\end{align}
where $\{x_1^\star,\ldots,x_k^\star\}
\subset
\{x_1,\ldots,x_N\}$ are the inputs corresponding to the $k$ nearest previously sampled outcomes to $g(x)$, measured relative to the denoised set $\big\{
\mu_{D}(x_1),\ldots,\mu_{D}(x_N)
\big\}$. 
This acquisition promotes candidates whose plausible outcomes are far from what has already been observed, while naturally accounting for posterior uncertainty.

The acquisition in \eqref{eq:ns-ts} does not directly optimize the reachability metric in \eqref{eq:reachability}. Instead, it is a continuous archive-distance novelty score used to choose inputs whose plausible outcomes are far from what has already been observed. This design is intentional. A direct reachability acquisition would require a fixed behavior map and would sum posterior probability mass over behavior regions not yet reached, as derived in the Supplementary Information (Section~1). Such an acquisition is useful when the behavior partition is known and computationally manageable, but it can be less convenient when the practitioner can define meaningful outcome distances before knowing the right behavior partition. BEACON therefore uses the distance-based novelty score as a general-purpose policy for improving reachability while avoiding direct dependence on the evaluation discretization.

To generate $g$ efficiently, we use the decoupled posterior sampling approach of~Wilson et al.\cite{wilson2020efficiently}, which yields accurate and differentiable approximate GP samples. This is particularly useful because it allows the resulting acquisition function to be optimized with gradient-based methods.
In other words, for each BEACON decision step, all random quantities defining the TS are drawn once and then held fixed while the acquisition function is optimized. Thus, the optimizer evaluates a deterministic sampled acquisition surface within that decision step.

We use Thompson sampling because it provides a simple pathwise mechanism for incorporating posterior uncertainty into the novelty score. Once a posterior sample of the input-to-outcome map is drawn, the same archive-distance novelty calculation can be applied regardless of the chosen outcome representation or distance metric. Alternative acquisitions, such as a UCB-style novelty score, are possible but would require additional modeling choices, tuning parameters, or numerical approximations. We do believe, however, that deriving and studying additional acquisition functions for NS is an interesting future work.

\subsection{Overview of the BEACON Algorithm}

\begin{figure}[t]
	\centering
	\includegraphics[width=\linewidth]{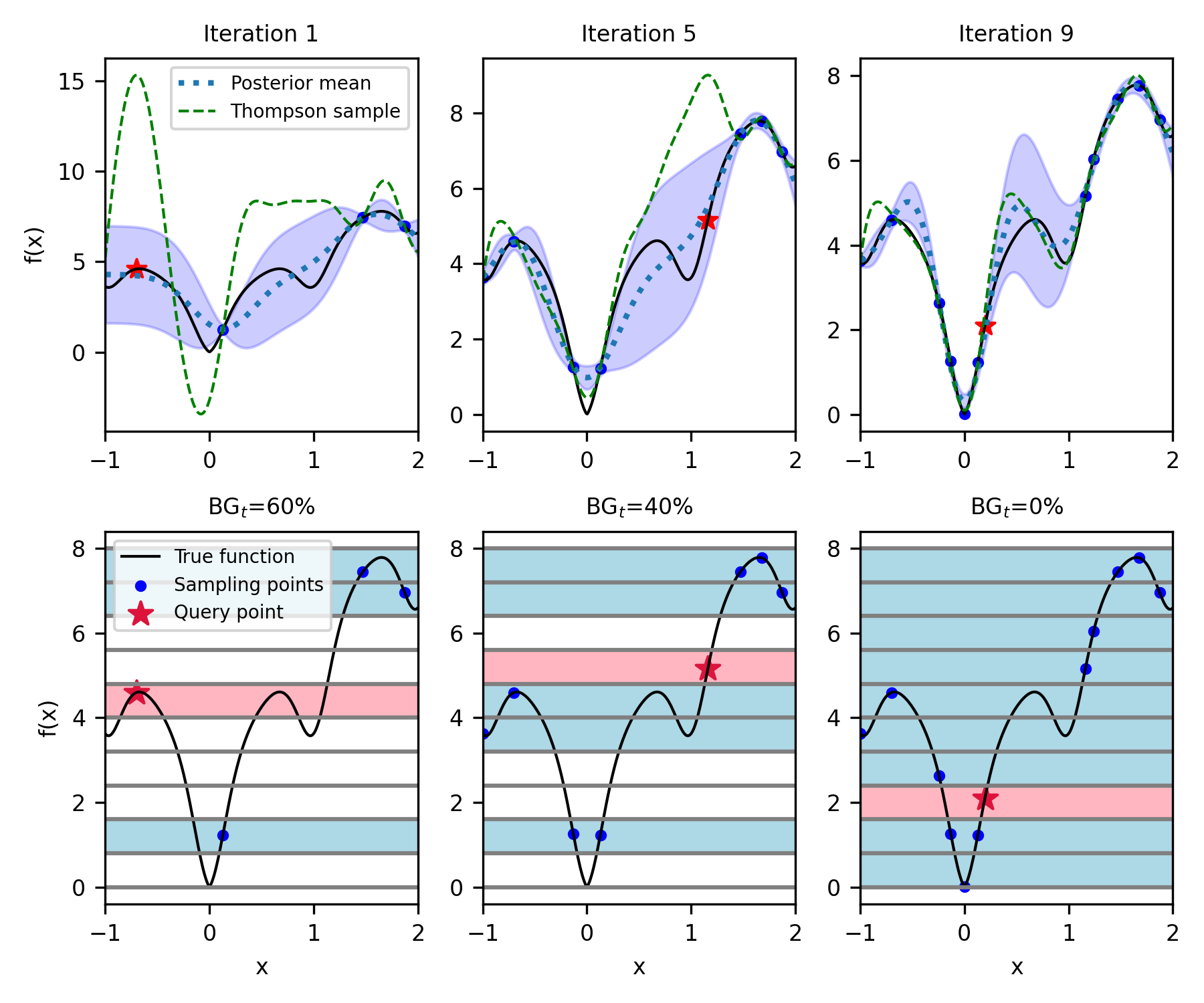}
	\caption{
		Visual illustration of BEACON on a one-dimensional test problem. Top: true function (black), GP posterior mean (blue dashed), and a Thompson sample (green dashed) at iterations 1, 5, and 9. Shaded regions indicate posterior uncertainty, and the red star marks the selected query point. Bottom: discretized behavior bins over the outcome space. Blue bands denote previously observed behaviors, while the red band denotes the predicted behavior of the next query. As BEACON progresses, it efficiently increases the reachability metric $\mathrm{Reach}_t$ by identifying previously unexplored regions.
	}
	\label{fig:beacon-illustration}
\end{figure}

Algorithm~\ref{alg:BEACON} summarizes the proposed BEACON procedure. At each iteration, BEACON fits a MOGP surrogate to the available data, draws a posterior sample, and selects the next query point by maximizing the novelty-based acquisition function in \eqref{eq:ns-ts}. The new observation is then added to the dataset, and the process repeats until the evaluation budget is exhausted. Figure~\ref{fig:beacon-illustration} provides a schematic illustration.

For clarity, Algorithm~\ref{alg:BEACON} is written in a standard sequential setting. However, because the method is built around posterior sampling, parallel and asynchronous variants can be developed using ideas from prior work on parallel (distributed) BO~\cite{kandasamy2018parallelised}.

\begin{algorithm}[t]
	\caption{BEACON}
	\label{alg:BEACON}
	\begin{algorithmic}[1]
		\State \textbf{Input:} Outcome function $f$, domain $X$, initial dataset $D_0$, total budget $T$, MOGP prior $MOGP(\mu,\kappa)$, neighbor count $k$, and outcome distance metric $\mathrm{dist}(\cdot)$
		\For{$t=1$ to $T$}
		\State Fit MOGP posterior: $MOGP_{t-1} \leftarrow f\mid D_{t-1}$
		\State Sample $g \sim MOGP_{t-1}$
		\State $x_t \leftarrow \argmax_{x\in X} \alpha_{\mathrm{NS}}(x\mid g,D_{t-1})$
		\State Observe outcome $y_t = f(x_t) + \eta_t$
		\State Update $D_t \leftarrow D_{t-1} \cup \{(x_t,y_t)\}$
		\EndFor
		\State \textbf{Return:} Final dataset $D_T$
	\end{algorithmic}
\end{algorithm}

\subsection{Gradient-Based Optimization of Acquisition Function}
\label{subsec:gradient-acq}

Efficiently maximizing $\alpha_{\mathrm{NS}}$ is essential to the practical performance of BEACON. For continuous input domains, we optimize the acquisition using gradient-based methods such as L-BFGS-B~\cite{byrd1995limited} with multiple random restarts.

Although the novelty score depends on a $k$-nearest-neighbor operation, it can be written in a form amenable to differentiation:
\begin{align}
\label{eq:ns-ts-sort}
\alpha_{\mathrm{NS}}(x\mid g,D)
=
\frac{1}{k}\,
e_k^\top
\,
\mathrm{sort}_{\uparrow}
\!\left(
\left\{
\mathrm{dist}\big(g(x),\mu_{D}(x_i)\big)
\right\}_{i=1}^N
\right),
\end{align}
where $e_k$ is a vector whose first $k$ entries are equal to $1$ and whose remaining entries are $0$, and $\mathrm{sort}_{\uparrow}(\cdot)$ denotes sorting in \emph{ascending} order. This ascending sort is necessary because novelty is defined using the $k$ \emph{nearest} previously observed outcomes. As discussed in Prillo et al.\cite{prillo2020softsort}, sorting operators admit continuous relaxations that are differentiable almost everywhere, which makes gradient-based optimization practical in this setting.

The cost of evaluating \eqref{eq:ns-ts-sort} scales linearly with $N$, the number of archived outcomes used in the nearest-neighbor calculation at the current decision step. In the experiments considered here, $N$ remains modest because the evaluation budgets are small and each evaluation is assumed to be expensive. For larger campaigns, the archive can be compressed by retaining representative outcomes rather than every previously sampled outcome, or by using standard nearest-neighbor data structures. For example, if several observations fall in the same behavior region or are very close in outcome space, one can retain a representative point for that local region of the archive. The effectiveness of this reduction is problem dependent. With fine behavior partitions or higher-dimensional outcome spaces, new evaluations may continue to occupy distinct regions, in which case the representative archive can grow nearly as fast as the total number of samples. Thus, archive management becomes more important as evaluation budgets increase, while remaining a secondary cost in the expensive-evaluation settings emphasized in this work. We do show that BEACON is able to effectively scale to several thousand data points in Section~5 of the Supplementary Information, especially when taking advantage of efficient large-scale GP implementations.

\subsection{Scaling BEACON to High-Dimensional Problems}

A major advantage of BEACON is that it can leverage modern advances in BO surrogate modeling. This is particularly important when the input dimension is large, since GP surrogates generally require additional structure to remain statistically and computationally effective. We consider two complementary strategies.

\subsubsection{SAAS Priors}

The sparse axis-aligned subspace (SAAS) prior~\cite{eriksson2021high} is a fully Bayesian approach that induces adaptive sparsity in GP models by placing a hierarchical prior over the input lengthscales. This encourages explanations that depend on only a small subset of the input variables, which is beneficial when outcome behavior is driven by a limited number of relevant features. Because of its generality, SAAS provides a useful default option for high-dimensional continuous design spaces. We use this approach in the high-dimensional LogD case study presented later.

\subsubsection{Chemistry-Aware GP Surrogates}

When the design variables are discrete, structured, or highly correlated (as is common for molecular representations), Euclidean distance-based kernels may be inadequate. In such cases, BEACON can be paired with chemistry-aware GP surrogates built from molecular kernels or structured representations. In our ESOL case study, for example, we use a Tanimoto-fragprint kernel motivated by the GAUCHE framework~\cite{griffiths2024gauche}. More broadly, BEACON is compatible with GP models defined over strings, fingerprints, graphs, and related molecular objects, provided posterior sampling is available; Section~2 of the Supplementary Information gives the exact surrogate choices used in each case study.

\subsection{User-Guided BEACON with Behavior Constraints}

In many applications, users have prior knowledge or preferences about which behaviors are worth exploring. For example, an experimentalist may wish to avoid regions that have already been sufficiently characterized or restrict attention to outcome regimes of greatest scientific interest. To incorporate such information, we introduce \emph{UG-BEACON}, a user-guided variant of BEACON that imposes constraints directly in behavior space.

Let $\mathcal{G} = \{G_1,\ldots,G_M\}$ denote a user-defined partition of the outcome space into non-overlapping regions. This partition can be tailored to the application and need not match the discretization used to compute reachability. Let $\mathcal{G}_{\mathrm{visited}} \subseteq \mathcal{G}$ denote the subset of regions that are either already explored or intentionally excluded by the user. UG-BEACON defines the feasible set
\begin{align*}
X_{\mathrm{feas}}
=
\left\{
x\in X
\; : \;
g(x) \notin \bigcup_{G\in\mathcal{G}_{\mathrm{visited}}} G
\right\},
\end{align*}
where $g$ is the current Thompson sample (TS) from the posterior model. The next evaluation is then chosen by solving
\begin{align*}
x^\star
=
\argmax_{x}
\ \alpha_{\mathrm{NS}}(x\mid g,D)
\quad
\text{subject to}
\quad
x\in X_{\mathrm{feas}} \cap X.
\end{align*}

This modification allows BEACON to avoid redundant sampling and direct exploration toward user-relevant parts of the outcome space, while leaving the underlying surrogate model and posterior-sampling machinery unchanged. 
UG-BEACON can be viewed as replacing full-range coverage with a targeted notion of coverage over behavior regions that satisfy user-defined preferences.
Practical examples and implementation details for constrained behavior discovery are provided in Section~3 of the Supplementary Information.

\section{Results and Discussion}
\label{sec:results-discussion}

We compare BEACON against several benchmark methods across synthetic, materials, molecular, reinforcement learning, and user-guided discovery tasks. Across these studies, the goal is behavior-space coverage under a limited evaluation budget. We therefore interpret the results as coverage experiments rather than standard scalar-objective optimization benchmarks.

Unless otherwise stated, all methods begin with 20 points sampled uniformly at random from $X$ to initialize the models. The baselines include evolutionary novelty search (NS-EA) \cite{lehman2011novelty}, novelty search based on distance to explored area (NS-DEA) \cite{doncieux2020novelty}, input-space novelty search (NS-FS), maximum posterior variance sampling (MaxVar), Sobol sequence sampling, and random search (RS), depending on which methods are applicable to the problem setting. Full implementation details are provided in Section~2 of the Supplementary Information. Each method then selects an additional 80-300 evaluations, depending on the problem. All experiments are repeated 20 times unless otherwise noted. Performance is measured using reachability, the fraction of predefined behavior regions discovered after $t$ evaluations. In all plots, we report the mean reachability together with $\pm$ one standard deviation across independent replicates. For problems defined over discrete candidate sets, the evolutionary NS baselines are not directly applicable, so comparisons are restricted to methods that can operate in that setting. 

Additional problem-specific details and ablation studies on the impact of the behavior space resolution, neighborhood size $k$, and observation noise can be found in the Supplementary Information (Section~4).
To demonstrate the flexibility of the user-guided extension, we also consider two constrained discovery tasks using UG-BEACON: an MNIST digit discovery problem and an oil sorbent material design problem.

\subsection{Synthetic Benchmark Problems}

We first consider scalar-output coverage landscapes adapted from the \textbf{Ackley}, \textbf{Rosenbrock}, and \textbf{Styblinski-Tang} functions. These functions are commonly used as global optimization benchmarks, but we do not use them here in that context. Instead, the scalar function value is treated as a one-dimensional outcome, and the goal is to cover the range of attainable outcome values. The outcome range is partitioned into 25 equally spaced intervals to define behavior regions. We study each function in three input dimensions, $d\in\{4,8,12\}$; complete function definitions/setup details are provided in Section~3 of the Supplementary Information.

Figure~\ref{fig:results-synthetic} summarizes the reachability results across these nine problems. BEACON consistently attains the highest reachability and often approaches complete coverage of the discretized behavior space within the allowed budget. The performance gap relative to the baselines generally widens as the input dimension increases, which is consistent with the expected difficulty of exploring high-dimensional spaces using purely input-space sampling or evolutionary heuristics.

To further test the multi-output setting, we also construct a synthetic \textbf{Multi-Output Plus} problem with a long-tailed two-dimensional outcome distribution. This function has six inputs and two outputs (full definition is in Section~3.1 of the Supplementary Information). Figure~\ref{fig:multi-output-plus-dist} (top) shows the distribution of outcomes induced by 10{,}000 uniformly sampled inputs from $X = [-5,5]^6$. Reachability is computed using a $10 \times 10$ grid over the two-dimensional outcome space.
This example provides a controlled vector-valued coverage problem in which the outcome distribution can be visualized directly. It is useful because the attainable outcomes are highly non-uniform, i.e., a large fraction of the probability mass is concentrated near the center, while many behavior regions lie in low-density tails.
This structure makes the problem challenging for NS because search procedures can easily oversample dense regions while missing the tails. As shown in Figure~\ref{fig:multi-output-plus-dist} (bottom), BEACON substantially improves coverage of the long-tail outcome regions, reaching nearly 80\% reachability, whereas competing methods remain below roughly 40\%.

\begin{figure*}[ht]
	\centering
	\includegraphics[width=0.8\textwidth]{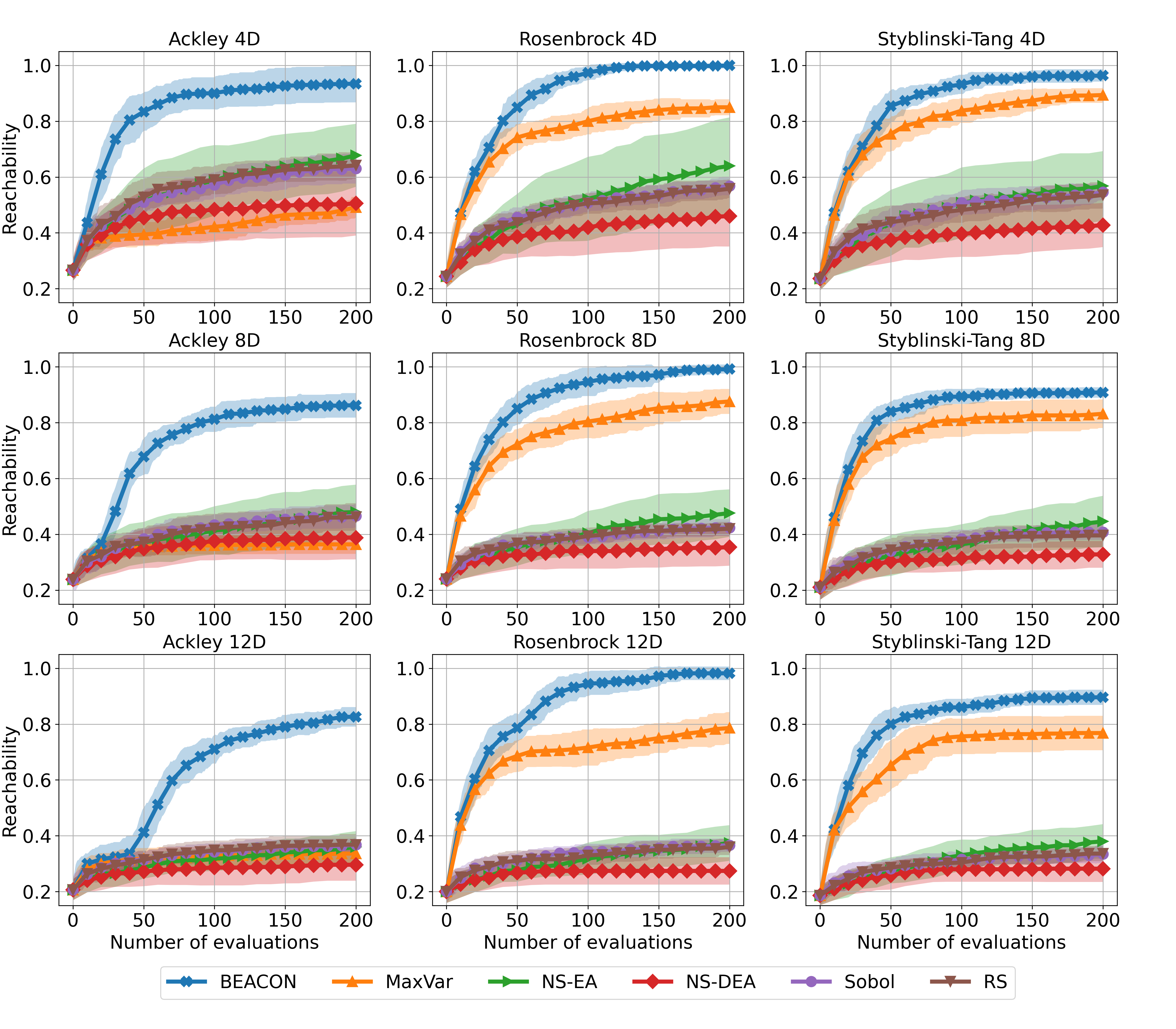}
	\caption{
		Reachability results on the synthetic benchmark problems: Ackley (left column), Rosenbrock (mid Dle column), and Styblinski-Tang (right column), with input dimensions 4 (top row), 8 (mid Dle row), and 12 (bottom row). Curves show the mean reachability and shaded bands indicate $\pm$ one standard deviation across 20 replicates. BEACON consistently achieves the strongest coverage of the outcome space, with particularly large gains in higher dimensions.
	}
	\label{fig:results-synthetic}
\end{figure*}

\begin{figure}[!ht]
	\centering
	\includegraphics[width=\columnwidth]{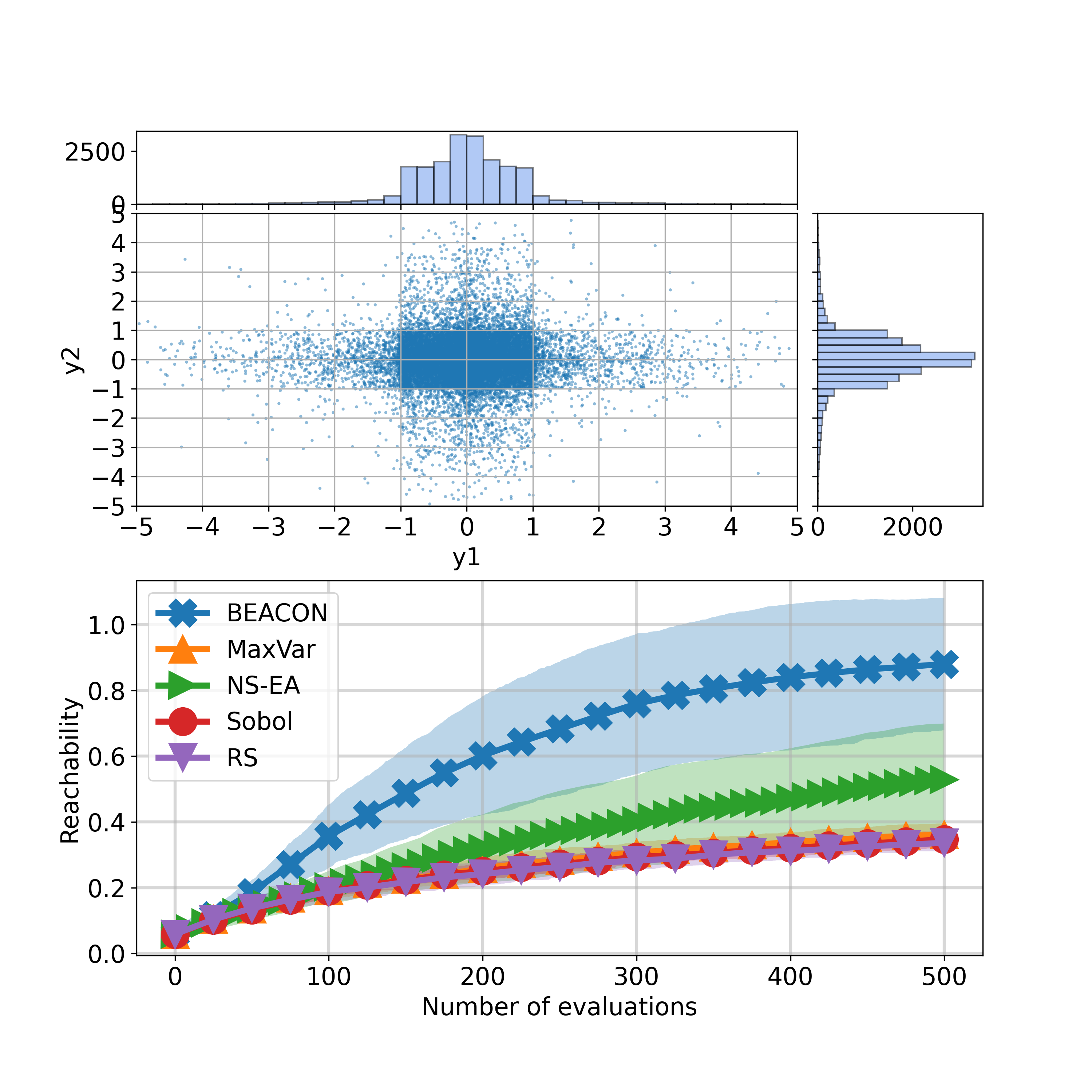}
	\caption{
		Results for the Multi-Output Plus test problem. Top: outcome distribution obtained from 10{,}000 uniformly sampled inputs, illustrating the strongly nonuniform and long-tailed structure of the two-dimensional behavior space. Bottom: reachability versus iteration for all methods. BEACON more effectively uncovers low-density outcome regions than the competing baselines.
	}
	\label{fig:multi-output-plus-dist}
\end{figure}

\subsection{Materials Discovery Applications}

We next consider discovery tasks involving metal-organic frameworks (MOFs), a chemically diverse class of porous crystalline materials relevant to gas storage and separations. Because MOFs span a vast combinatorial design space over linkers, metal centers, topologies, and defects, it remains difficult to characterize the full diversity of properties that can arise in practice~\cite{lee2021computational}. Existing MOF databases are also known to exhibit biases that can affect downstream screening and learning studies~\cite{moosavi2020understanding}. These features make MOFs a natural setting for novelty-driven search, especially when evaluations are computationally or experimentally expensive.

In each task, a candidate MOF is represented by a fixed-dimensional descriptor vector $x \in X \subset \mathbb{R}^d$, and the property or properties of interest define the outcome $y = f(x) \in O$. Descriptor definitions and outcome distributions are provided in Section~3 of the Supplementary Information.

\paragraph*{Hydrogen uptake capacity.}
We first consider a hydrogen storage dataset from~\cite{ghude2023exploring} containing 98{,}000 unique MOFs. Each MOF is represented by 7 real-valued descriptors, and the outcome is a scalar hydrogen uptake value.

\paragraph*{Nitrogen uptake capacity.}
The second task considers nitrogen uptake in MOFs for gas separation, using the dataset from~\cite{daglar2022combining}. This dataset contains 5{,}224 MOFs, each represented by 20 descriptors, with scalar nitrogen uptake as the outcome.

\paragraph*{Joint CO$_2$ and CH$_4$ uptake.}
The third task considers simultaneous exploration of carbon dioxide and methane uptake capacities using the dataset from Moosavi et al.~\cite{moosavi2020understanding} with 7{,}000 MOFs. Here, the input consists of 25 descriptors and the outcome is two-dimensional. The corresponding outcome distribution is strongly skewed; see Section~3 of the Supplementary Information.

Figure~\ref{fig:results-MOF} compares BEACON with the applicable baselines on these three tasks. Because the candidate MOFs come from discrete datasets, the evolutionary NS baselines are not included here. Across all three problems, BEACON achieves the strongest reachability and, within the available sampling budget, attains full coverage of the discretized behavior space. These results highlight the value of sample-efficient NS for property-space exploration in chemically large but evaluation-limited settings.

\begin{figure}[!t]
	\centering
	\includegraphics[width=\columnwidth]{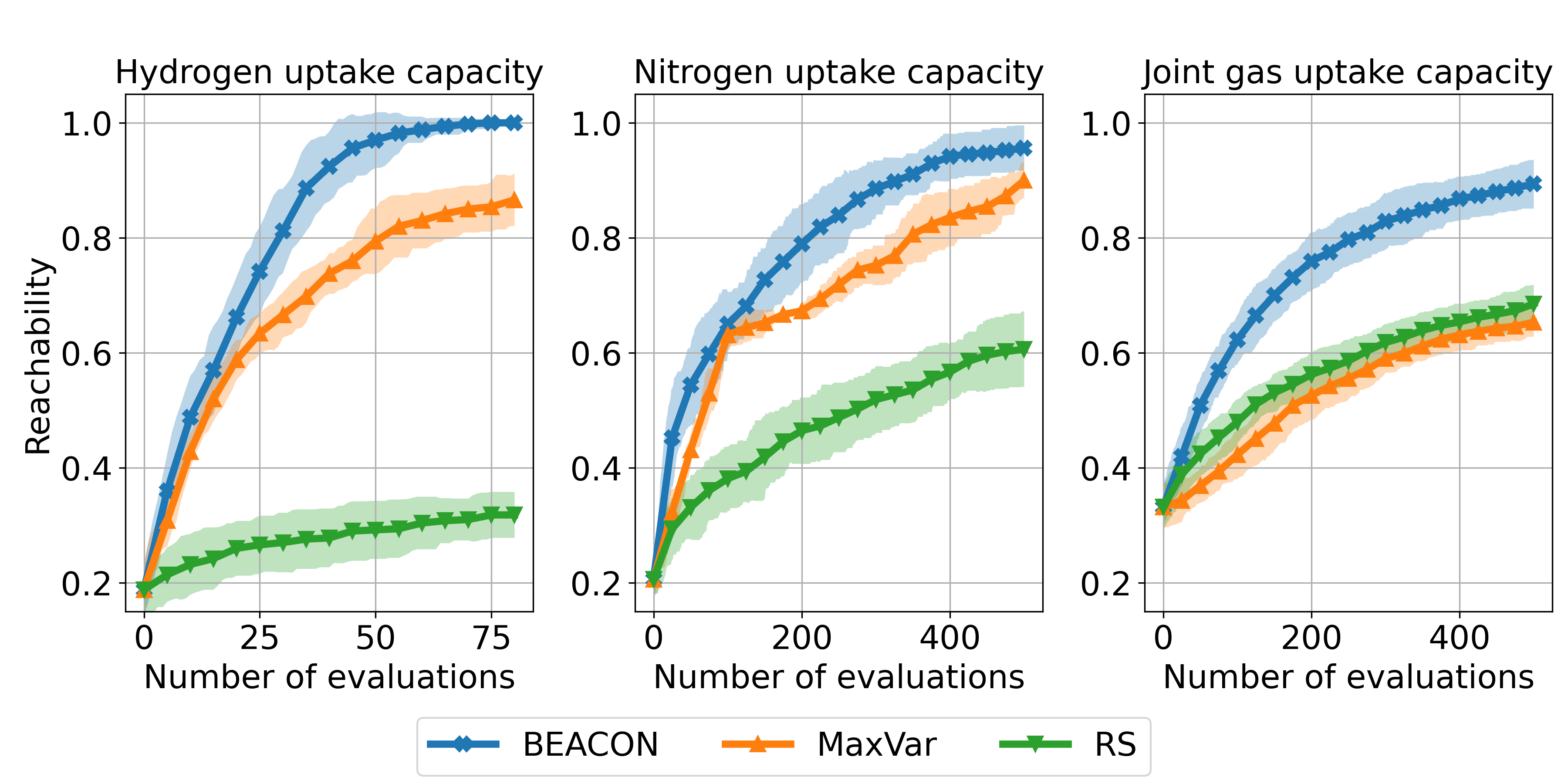}
	\caption{
		Reachability results for the MOF discovery tasks: hydrogen uptake (left), nitrogen uptake (mid Dle), and joint CO$_2$/CH$_4$ uptake (right). Because these problems are defined over discrete candidate sets, only methods applicable in that setting are shown. BEACON achieves the highest reachability in all three cases and reaches full coverage within the allowed budget.
	}
	\label{fig:results-MOF}
\end{figure}

\subsection{Molecular Discovery Applications}
\label{subsec:solubility}

We next study molecular property discovery problems relevant to pharmaceutical and chemical design. In these settings, the objective is not necessarily to identify a single molecule with the best property value, but rather to discover molecules spanning a broad range of property behaviors. Such diversity can support model development, dataset enrichment, and early-stage screening. 
Across all tasks, the input space $X\subset\mathbb{R}^d$ encodes molecular structure using descriptor-based or high-dimensional molecular representations, while the outcome is a scalar molecular property. These examples thus test BEACON in high-dimensional molecular input spaces with scalar behavior-space coverage metrics.

\paragraph*{Water solubility.}
We use the dataset from Boobier et al.~\cite{boobier2020machine} containing 900 small organic molecules, each represented by a 14-dimensional descriptor vector.

\paragraph*{ESOL.}
We next consider the ESOL dataset~\cite{delaney2004esol}, which contains 1{,}128 molecules with experimentally measured aqueous solubility. Each molecule is represented by a 2{,}133-dimensional binary fragprint vector. For this task, BEACON uses a Tanimoto-fragprint kernel motivated by chemistry-aware GP modeling~\cite{griffiths2024gauche}.

\paragraph*{LogD.}
Finally, we consider a LogD dataset containing 2{,}070 molecules from Win et al.~\cite{win2023using}, each represented by 125 features. To address the higher dimensionality, we pair BEACON with the sparse axis-aligned subspace (SAAS) prior~\cite{eriksson2021high}.

In these experiments, reachability is computed over the full observed property range to provide a neutral comparison of broad property-space coverage. For molecular campaigns where the scientific goal is to identify many candidates within a high-performing region, the behavior space can instead be restricted to a preferred property window, weighted toward high-value regions, or discretized more finely in those regions. The user-guided formulations in Section \ref{subsec:user-guide-examples} illustrate how BEACON can be adapted to this targeted-coverage setting.

\begin{figure}[!t]
	\centering
	\includegraphics[clip=true,width=\columnwidth]{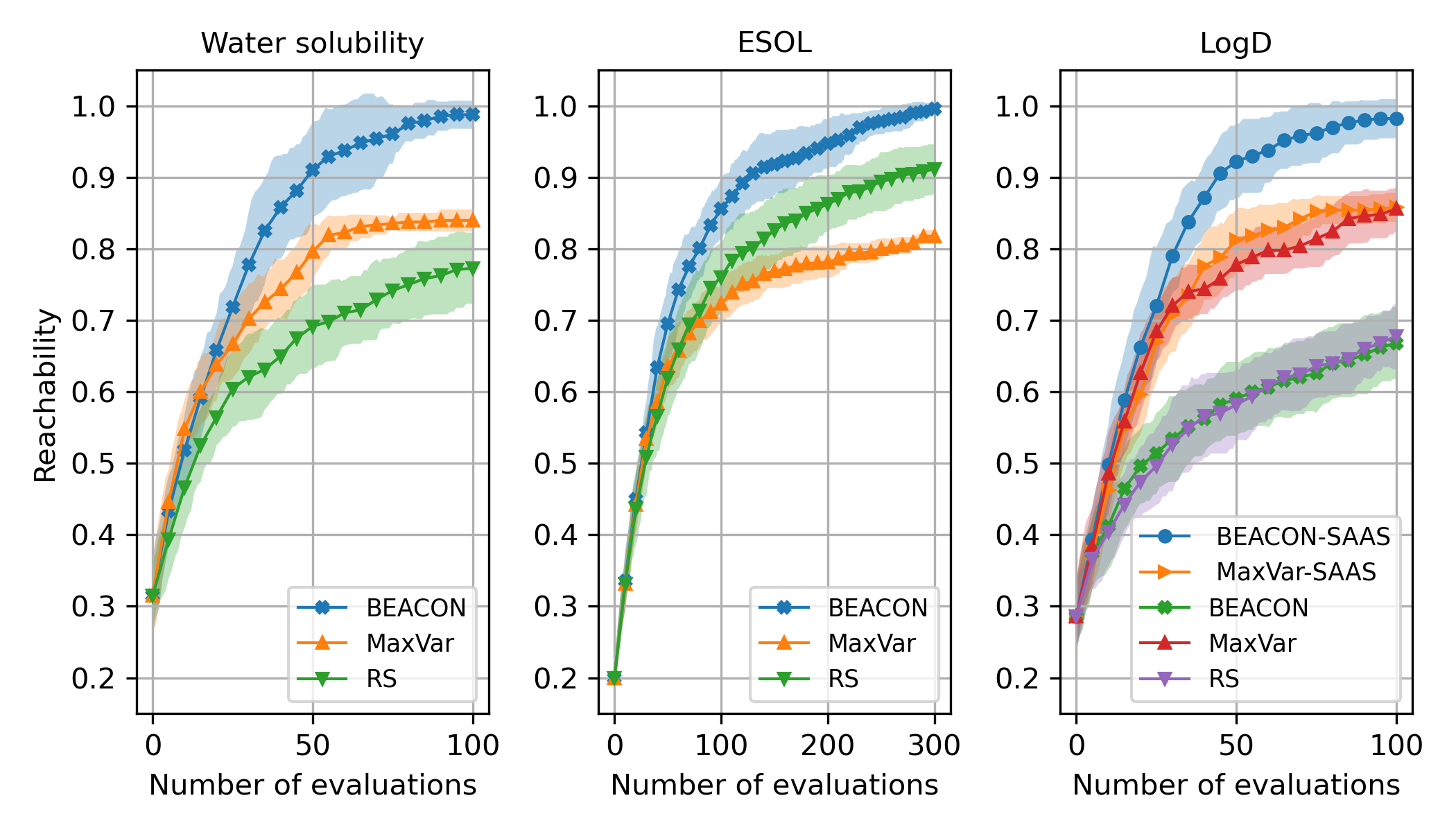}
	\caption{
		Reachability results for the molecular discovery tasks: water solubility (left), ESOL (mid Dle), and LogD (right). BEACON consistently outperforms the benchmark methods, with especially strong gains in the higher-dimensional ESOL and LogD settings.
	}
	\label{fig:results-smallmolecule}
\end{figure}

The results in Figure~\ref{fig:results-smallmolecule} show that BEACON consistently achieves the highest reachability across all three tasks. The gains are especially pronounced in ESOL and LogD, where the use of task-appropriate surrogate structure is important for effective search. Overall, these results support the view that BEACON can be naturally combined with chemistry-aware GP models to improve sample-efficient exploration of molecular property space.

\subsection{A Deceptive Reinforcement Learning Landscape}

We next consider a reinforcement learning (RL) maze navigation task with a deliberately deceptive reward landscape. The goal is to control a ball from a fixed start location to a goal location within 300 time steps using a linear control policy with 8 tunable parameters. In this problem, improving reward locally does not necessarily correspond to exploring behaviorally useful parts of the maze; a reward-driven search can therefore become trapped by policies that make partial progress but fail to discover trajectories that reach the goal.

For BEACON and the NS baselines, the behavior outcome is the two-dimensional terminal position of the agent after executing the policy, and novelty is computed in this terminal-state space. Thus, these methods do not directly optimize reward. Instead, they search for policies that lead to new terminal states, which can help overcome the deceptive reward structure by encouraging broader exploration of the maze. In contrast, expected improvement (EI) \cite{jones1998efficient} is included as a reward-driven scalar-objective comparator; its GP surrogate models normalized reward as a function of the controller parameters, and its acquisition seeks policies with high expected reward improvement. We report normalized reward separately as a task-success metric because it measures progress toward the goal and enables comparison with this reward-driven baseline. A schematic of the maze and additional details are given in Section~3.4 of the Supplementary Information.

As shown in Figure~\ref{fig:maze-result}, BEACON substantially outperforms both the novelty-based baselines and the reward-driven methods, including EI. The top panel reports the average best observed reward over 20 replicates, while the bottom panel shows the distribution of final best rewards. BEACON is the only method that solves the task across all replicates, demonstrating that sample-efficient NS can be particularly valuable when objective-based search is vulnerable to deception.

\begin{figure}[!ht]
	\centering
	\includegraphics[width=1\columnwidth,clip=true]{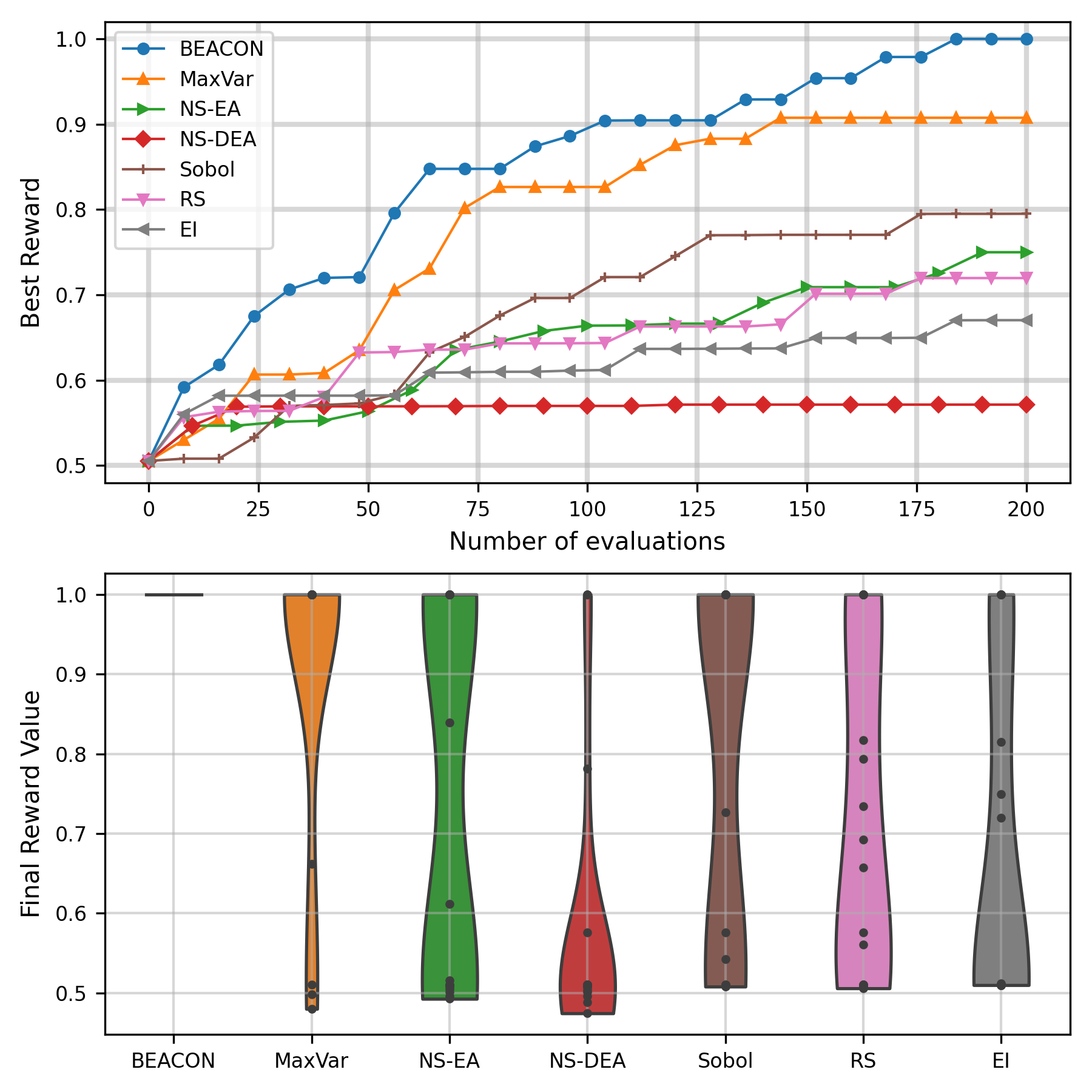}
	\caption{
		Results for the RL maze-navigation case study. Top: average best observed reward versus iteration for all methods across 20 replicates. Bottom: violin plot of the best reward obtained at the final iteration. BEACON is the only method that consistently achieves successful maze completion across all replicates.
	}
	\label{fig:maze-result}
\end{figure}

\subsection{User-Guided Discovery with Behavior Constraints}
\label{subsec:user-guide-examples}

We finally evaluate UG-BEACON on two tasks where user-defined behavior preferences are incorporated directly into the search.

\begin{figure}[!ht]
	\centering
	\includegraphics[width=1\columnwidth,clip=true]{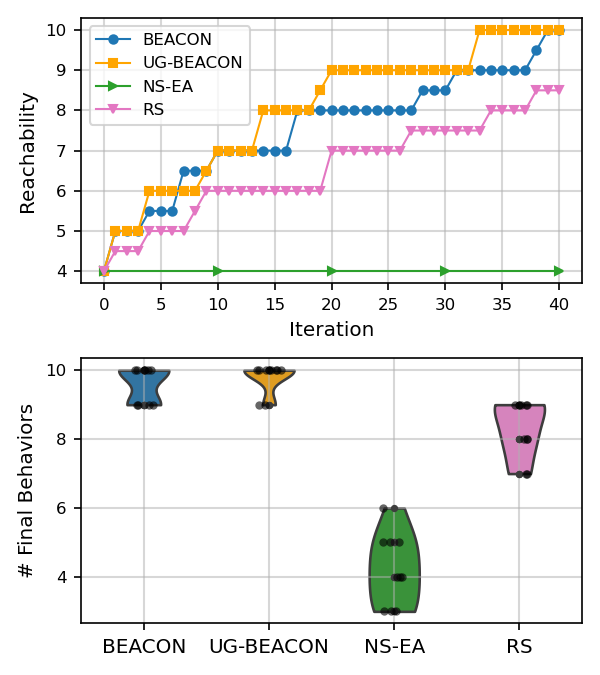}
	\caption{
		Results for the MNIST digit discovery case study. Top: average number of unique digit classes discovered versus iteration across 10 replicates. Bottom: distribution of the number of discovered classes at the final iteration. UG-BEACON reaches full class coverage more rapidly by incorporating behavior-level constraints based on previously observed digits.
	}
	\label{fig:MNIST-result}
\end{figure}

\paragraph*{MNIST digit discovery.}
The first example uses the MNIST dataset~\cite{deng2012mnist}. The goal is to discover as many distinct handwritten digit classes (0-9) as possible by sampling from the latent space of a trained variational autoencoder (VAE)~\cite{doersch2016tutorial}. Each sampled latent point is decoded to a $14\times14$ image, which is treated as the black-box outcome. Thus, BEACON computes novelty in the 196-dimensional pixel-output space (stacked $14 \times 14$ outputs into a vector) rather than directly in digit-label space. A convolutional neural network (CNN) classifier is then used to assign each generated image to a digit class. UG-BEACON uses these class labels to avoid revisiting already discovered digits during the search, and reachability is computed over the ten digit classes. As shown in Figure~\ref{fig:MNIST-result}, UG-BEACON discovers the full set of digit classes more rapidly than both BEACON and the baseline methods.

\begin{figure}[ht]
\centering
\includegraphics[width=0.49\columnwidth,clip=true]{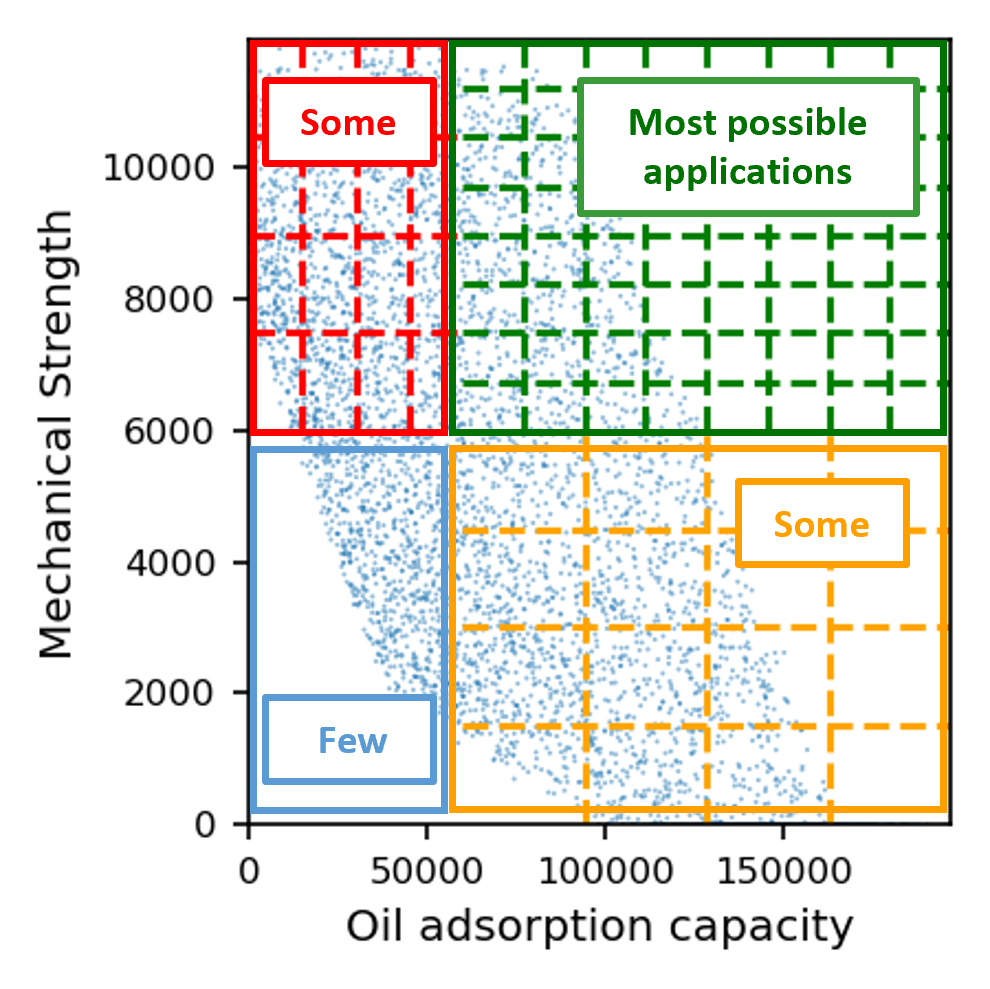}
\includegraphics[width=0.49\columnwidth,clip=true]{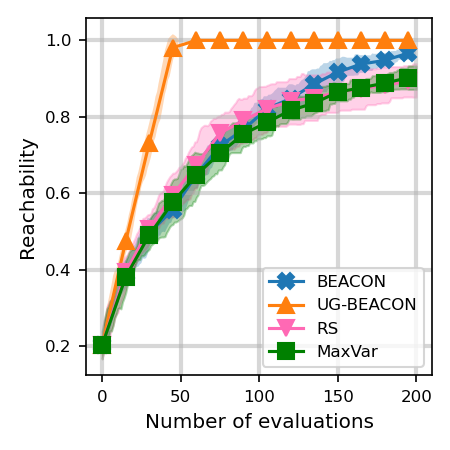}
\caption{
	Oil sorbent material discovery case study. Left: user-defined partition of the two-dimensional outcome space, with finer discretization in the most desirable performance regions. Right: reachability versus iteration for UG-BEACON and the benchmark methods. UG-BEACON attains full preferred-space coverage substantially faster than the competing approaches.
}
\label{fig:oil-sorbent}
\end{figure}

\paragraph*{Oil sorbent materials.}
As a second example, we consider the discovery of oil sorbent materials under application-driven outcome preferences. Here, the two-dimensional outcome space is partitioned non-uniformly according to utility, i.e., regions corresponding to high adsorption capacity and high strength are discretized more finely, while less relevant regions are coarsened. This allows the user to place greater emphasis on discovery in scientifically/practically valuable regimes. UG-BEACON uses this partition both to focus the search and to avoid redundant sampling of already explored regions. As shown in Figure~\ref{fig:oil-sorbent}, UG-BEACON quickly reaches full coverage of the preferred outcome space, outperforming BEACON and the competing baselines.

Taken together, these examples show that UG-BEACON can incorporate user guidance in a simple and effective way, making the framework useful not only for unconstrained NS, but also for targeted discovery tasks in which some outcome regions are more important than others.

\subsection{Practical considerations for sharp outcome spaces}

A practical limitation of BEACON, and more broadly of GP-guided sequential design methods, is that discovery depends on the surrogate model being able to represent the relevant input-outcome structure. Thompson sampling encourages exploration of uncertain regions, but it cannot guarantee discovery of an arbitrarily small or highly localized behavior region if the initial data provide no signal that such a region exists. This issue is relevant in materials discovery problems with, e.g., phase changes or other significant/sharp changes in mechanism. 

In the most difficult case, the response may appear nearly flat over most of the input space but contain abrupt, weakly structured jumps. A standard stationary GP can easily over-smooth such behavior, and posterior samples from a misspecified model may not assign enough probability to the rare region to guide sampling there. More flexible surrogates can help when the sharp structure is learnable from the available data. Examples include non-stationary kernels or input warping for spatially varying length scales~\cite{snoek2014input}, adaptive hyperparameter treatments that reduce the risk of over-smoothing~\cite{berkenkamp2019no}, deep kernel learning models that learn representations jointly with the GP~\cite{wilson2016stochastic}, and methods designed to account for GP model misspecification~\cite{bogunovic2021misspecified}. These approaches improve the modeling side of the problem, but they cannot overcome the absence of learnable structure. If a rare behavior is effectively uncorrelated with nearby sampled points and no available side information indicates where it may occur, then no GP-guided acquisition can reliably target it from a small number of evaluations; discovery is possible only through sufficiently broad exploration or chance sampling.

For prospective applications where missing such regions would be costly, BEACON can be combined with more deliberate exploration safeguards. These include broader space-filling initialization, maintaining an input-space diversity component, or adding scheduled exploration when reachability begins to plateau. Such modifications are directly compatible with the framework because they change the surrogate model or the candidate-generation strategy while preserving the core idea of scoring novelty in outcome space. Another potential strategy is to iteratively shrink the search region around the most novel previously discovered outcomes, thereby progressively increasing the resolution of the surrogate model in the sharp regions likely to contain highly novel outcomes ~\cite{siemenn2023fast}. Developing these hybrid strategies is an important direction for future work, particularly for discovery problems where broad coverage must be balanced against reliable identification of rare high-performing regions.

\section{Conclusions}
\label{sec:conclusions}

In this work, we presented BEACON, a Bayesian novelty search algorithm for discovering diverse behaviors of expensive, noisy black-box systems. By combining multi-output Gaussian process surrogates with a Thompson sampling-based novelty acquisition function, BEACON directs evaluations toward previously unexplored regions of outcome space while still accounting for model uncertainty and observation noise. This means that BEACON is particularly well suited to low-data discovery settings, where evaluations are costly and a broad coverage of outcomes is more valuable than optimizing a single predefined target.

Across synthetic benchmarks and real-world case studies in materials and molecular discovery, BEACON consistently achieved stronger reachability than competing methods under limited evaluation budgets. These results suggest that Bayesian surrogate modeling can make novelty-driven exploration practical in domains where traditional evolutionary NS approaches are too sample-inefficient to be useful.

There are several important directions for future work. From a methods standpoint, reducing computational overhead and developing stronger theoretical guarantees remain worthwhile challenges. From an applications standpoint, a particularly promising next step is to integrate BEACON-like strategies into closed-loop autonomous experimentation platforms, where the goal is not only to optimize a known target but also to uncover previously unseen phenomena or property regimes~\cite{bulanadi2025beyond}. We view this as an especially compelling opportunity for truly autonomous discovery in chemistry and materials science.

\section*{Author contributions}
Wei-Ting Tang: Conceptualization, Methodology, Software, Formal analysis, Investigation, Visualization, Writing – original draft.
Ankush Chakrabarty: Methodology, Software, Formal analysis, Writing – review $\&$ editing.
Joel A. Paulson: Conceptualization, Supervision, Funding acquisition, Project administration, Writing – review $\&$ editing.

\section*{Conflicts of interest}
There are no conflicts to declare

\section*{Data availability}
The code for BEACON is publicly available at \url{https://github.com/PaulsonLab/BEACON} and archived on Zenodo (\url{https://doi.org/10.5281/zenodo.20435036}). The repository includes the main scripts used to run BEACON in continuous and discrete settings for both single- and multi-outcome problems, together with the Thompson sampling implementation used throughout the study. It also contains problem-specific code and supporting folders for the synthetic, materials, molecular, and MNIST case studies, as well as plotting utilities. These resources are intended to support reproduction of the empirical results reported in this paper and to provide a starting point for applying BEACON to related novelty search (NS) problems. Datasets for molecular and material problems studied in this work are available at \url{https://doi.org/10.6084/m9.figshare.32493084}.

\section*{Acknowledgments}
This work was supported by the National Science Foundation (NSF) CAREER Award 2237616.



\balance


\bibliography{rsc} 
\bibliographystyle{rsc} 

\clearpage
\includepdf[pages=-]{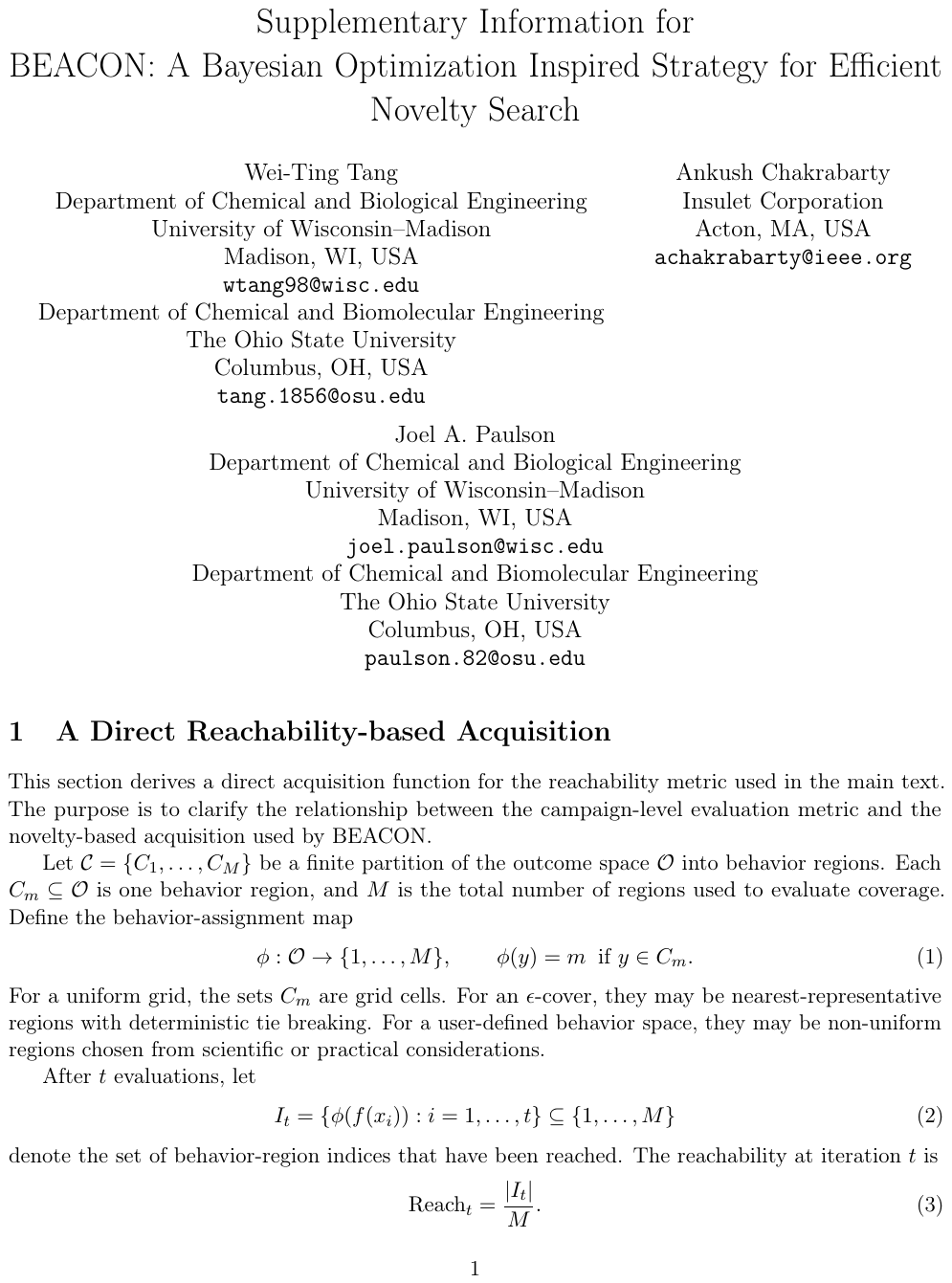}
\end{document}